\documentclass[manuscript,screen]{acmart}

\AtBeginDocument{%
  \providecommand\BibTeX{{%
    \normalfont B\kern-0.5em{\scshape i\kern-0.25em b}\kern-0.8em\TeX}}}

\setcopyright{acmlicensed}
\copyrightyear{2018}
\acmYear{2024}
\acmDOI{XXXXXXX.XXXXXXX}

\acmConference[Conference acronym 'XX]{Make sure to enter the correct
  conference title from your rights confirmation emai}{June 03--05,
  2018}{Woodstock, NY}
\acmISBN{978-1-4503-XXXX-X/18/06}
\usepackage{placeins}

\usepackage[normalem]{ulem}

\begin{document}

\title{Causality extraction from medical text using Large Language Models (LLMs)}

\author{Seethalakshmi Gopalakrishnan}
\email{sgopala4@uncc.edu}
\orcid{0009-0006-8331-3476}
\affiliation{%
  \institution{University of North Carolina at Charlotte}
  \streetaddress{9201 University City Blvd}
  \city{Charlotte}
  \state{North Carolina}
  \country{USA}
  \postcode{28223}
}

\author{Luciana Garbayo}
\affiliation{%
  \institution{University of Central Florida}
  \streetaddress{6850 Lake Nona Blvd.}
  \city{Orlando}
  \state{Florida}
  \country{USA}}
\email{Luciana.Garbayo@ucf.edu}

\author{Wlodek Zadrozny}
\affiliation{%
  \institution{University of North Carolina at Charlotte}
  \streetaddress{9201 University City Blvd}
  \city{Charlotte}
  \state{North Carolina}
  \country{USA}}
\email{wzadrozn@uncc.edu}







\renewcommand{\shortauthors}{Gopalakrishnan, Garbayo and Zadrozny}

\begin{abstract}
This study explores the potential of natural language models, including  large language models, to  extract causal relations from medical texts, specifically from Clinical Practice Guidelines (CPGs). 
The outcomes  causality extraction from Clinical Practice Guidelines for gestational diabetes are presented, marking a first in the field. 
We report on 
a set of experiments using variants of BERT (BioBERT, DistilBERT, and BERT) and using Large Language Models (LLMs), namely GPT-4 and LLAMA2. 
Our experiments show that BioBERT performed better than other models, including the Large Language Models, with an average F1-score of 0.72.
GPT-4 and LLAMA2 results show similar performance but less consistency. 
We also release the code and an annotated a corpus of causal statements within the Clinical Practice Guidelines for gestational diabetes. 

\end{abstract}




\maketitle

\section{Introduction}
Clinical Practice Guidelines (CPGs) are a set of expert guidelines developed to guide physicians in navigating the complexities of the medical decision-making process. Various medical societies provide numerous such guidelines, based on their focus (e.g. cardiology vs. family medicine). This variability can lead to inconsistencies in the comparison and application of guidelines, as noted in our previous work \cite{hematialam2020computing, hematialam2021method}. Recognizing these discrepancies is crucial for effective communication between patients and physicians.  

Pre-trained language models like BERT \cite{devlin2018bert}, which dynamically adjusts the weightings between each part of the output and all elements of the input based on their connection (attention), have demonstrated remarkable effectiveness
on numerous natural language processing tasks \cite{devlin2018bert}, including causality extraction \cite{gopalakrishnan2023text}. 

More recent improvements come from Large Language Models (LLMs) like GPT-4 \cite{openai2023gpt4}, which are pre-trained on extensive data and later enhanced through reinforcement learning feedback from both humans and AI to ensure adherence with human principles and policy compliance. 
Another recent model, also used in this article, 
is the open source LLAMA2 \cite{touvron2023llama}. It was trained on 2 trillion tokens and it available in three different sizes (7B, 13B, and 70B). 
It is widely used in numerous tasks, particularly in information extraction \cite{wiest2023text}. And perhaps more importantly, LLAMA is a focus of research on understanding capabilities and structures in large language models \cite{chen2023beyond}, \cite{gurnee2023language}.


Despite the emergence of the LLMs, BERT continues to be one of the top-performing models for various applications, including causality extraction \cite{khetan2020causal}, \cite{lyu2022dcu},\cite{gopalakrishnan2023text}, \cite{peng2019transfer}. This study aims to  extract causal relations ('causalities') from the medical text in  Clinical Practice Guidelines.

In medicine, the best explanations are causal (and imply the opportunity for better recommendations?). Causal reasoning is (therefore) used (not only in producing and (but in) evaluating the impact of medical guidelines (over time in learning system loops).  The mechanistic model of (explanation of biological phenomena) is preferred in biomedicine (even if it is  probabilistic). 
Automated analysis is necessary (
for various applications like comparing differences in guidelines, diagnostic support, etc.)
since there are currently over \textit{37,000} of medical guidelines indexed on PubMed as "practice guidelines" and two orders of magnitude of articles that are used to produce the guidelines. {\textit{Most of them use causal statements}. }
The main contributions of this study are
\begin{itemize}
    \item An entirely new type of public dataset of cause/effect relationships for Clinical Practice Guidelines. For medical text, there are only a very few causality extraction datasets available \cite{mihuailua2013biocause}, \cite{reklos2022medicause}, but none of them focus on Clinical Practice Guidelines (CPGs). 
    
    \item A performance evaluation of several known Large Language Models (LLMs) on the corpus of CPGs for causality extraction task. The results indicate that the performance of GPT-4 does not increase with an increase in the prompt size beyond 10. The LLAMA2 performance does not improve with the increase in the number of epochs.
\end{itemize}

From the experiments and evaluation we conclude that 
variants of BERT might still be preferred for this task, given the ease of fine-tuning and consistent performance. With BERT, we obtained an average F1 score of 72\%, whereas GPT-4 gave an average F1 score of 60\%.  LLAMA2 shows promise, in that an average F1 score of 76\% was obtained on subset for which it made predictions; LLAMA2 did not generate predictions for 20-35+\% of data.  

\section{Related Work}
Causality extraction is the task of automatically extracting the cause/effect relationships from the text. In this section, we briefly discuss studies related to causality extraction. 

\subsection{Work related to automatic information extraction from Clinical Practice Guidelines (CPG)}
This section summarizes the prior work related to information extraction on the Clinical Practice Guidelines.
Extracting clinical findings from notes of outpatient progress was early done realized by \cite{ertle1996automated}. 
Fifteen years later \cite{taboada2013combining} targeted the automated extraction of diagnosis and treatment procedures from clinical guidelines. A similar work \cite{kaiser2010supporting} introduced a method for automatically collecting useful information using rules rooted in both syntactic and semantic information. A pattern-based approach was further used by \cite{chunhua2014towards}, which contrasted a manually developed ontology for CPG eligibility criteria with a top-level ontology stemming from a semantic pattern-based approach. A more recent work \cite{fazlic2019novel} introduced an innovative system that blends together the methodologies of Natural Language Processing (NLP) and Fuzzy Logic. A supervised machine learning methodology was used by another similar work \cite{graham2022associations} to extract and categorize Conflicts Of Interest (COIs) from disclosure statements indexed in PubMed.

Recently, Large Language Models (LLMs) have (also) been employed for a variety of NLP tasks, including those involving information extraction. LLMs can be fine-tuned to cater to a specific dataset, or a prompt-based approach can be utilized. An illustrative study \cite{zhao2021calibrate} measures the efficiency of the few-shot learning performance of GPT-3 in tasks related to text classification and information extraction. A recent survey article \cite{landolsi2023information} provides a summary of the methods and solutions employed for information extraction. It also highlights the challenges encountered when extracting information from medical documents (such as, ambiguities when the named entity belongs to more than one class, phrase boundary detection, name variations, and others).

\subsection{Recent work on causality extraction from non-medical text}
The study by \cite{li2021causality} directly extracts cause and effect from text without separately extracting candidate pairs and their relations. A work on event extraction \cite{man2022event} focuses on identifying the causal relationship between pairs of event mentions, also known as 'Event Causality Identification' (ECI). Balashankar et al. \cite{balashankar2019identifying} propose an event extraction (modality) that seeks to uncover the hidden relationships between events mentioned in news streams by creating a Predictive Causal Graph (PCG). Prompt tuning has been proposed to bridge the gap between pre-training and fine-tuning on many of the mainstream NLP tasks like text classification\cite{schick2020exploiting,zhang2021differentiable}, information extraction\cite{chen2021lightner, cui2021template} etc. In  \cite{liu2023kept}, Knowledge Enhanced Prompt
Tuning (KEPT) employs external knowledge sourced from knowledge bases (KBs) to fine-tune pre-trained language models through the design of an attention mechanism.

A recent article \cite{chan2023chatgpt} describes the use of ChatGPT to extract cause/effect relationships from text on three datasets: (1) Choice of Plausible Alternatives (COPA) \cite{gordon2012semeval}, which is a collection of premises, along with two questions related to each premise, that requires causal reasoning in order to solve the inference; (2) e-CARE \cite{du2022care} which is an explainable causal reasoning dataset with cause, effect and two possible explanations; (3) Headline Cause \cite{gusev2021headlinecause} dataset, which aims to identify the implicit causal relations between pair of text. On COPA, ChatGPT in-context learning got a 97\% accuracy (performance); on the eCARE dataset, a 79.6\% accuracy was obtained using prompt engineering, and 72.7\% accuracy was recorded for on Headline Cause. 




\subsection{Causality extraction from the medical text}
In 2013 the task of automatic detection conditional statements in medical guidelines was first introduced  \cite{wenzina2013identifying}. The article used a rule-based approach, focusing on presence of connectives such as "if", and a collection of word-based syntactic patterns. Subsequent works on detecting condition action statements from CPGs, \cite{hematialam2017identifying} and \cite{hematialam2021knowledge}, apply supervised machine learning techniques to classify sentences according to whether they express conditions and actions. Another study  \cite{hussain2018recommendation}  used heuristic patterns to identify recommendation statements in Clinical Practice Guidelines (CPG). A review article \cite{fu2020clinical} documents the existing methods and tools for clinical concept extraction. The summarization of biomedical literature is addressed in \cite{xie2022pre} using pre-trained language models.
A more recent study \cite{tang2023does} explores the use of ChatGPT for clinical text mining, specifically for extracting structured data from unstructured healthcare texts and focusing on biological named entity recognition and relation extraction by identifying and extracting medical entities from text related to disease and drug, symptoms and treatment, etc.

\section{Data}
We annotated seven documents of gestational diabetes (clinical practice) guidelines from various societies (and medical entities) like(such as) the American Diabetes Association (ADA) (\cite{american20202},\cite{metzger2010international}), US Preventive Services Task Force (USPSTF) (\cite{davidson2021screening}, \cite{pillay2021screening}), American College of Obstetrics \& Gynecology (ACOG) \cite{mellitus2018acog}, American Academy of Family Physician (AAFP) \cite{mills2021screening}, and Endocrine Society \cite{blumer2013diabetes}.

The decision to annotate gestational diabetes clinical practice guidelines was based on the opportunity to explore causality inference in the future with situated learning models with prediction (team member Dr. Garbayo worked on safety and quality database development on maternal and child in maternities, resulting in the creation of the largest maternity database in Latin America \cite{leal2004factors}.

Two annotators were recruited. Given a (medical) text document, their task was to read the document and mark the cause, effect, condition, action, modal, and degree of influence with tags. The cause was marked as C, effect as E, condition as CO, and action as A. The phrases containing any of these causal phrases should be differentiated; for example, the beginning of a cause phrase will be marked as <C> and the end as </C>.  For example:
\begin{example}
\texttt{
    <C>Pregnant persons with gestational diabetes</C> are at <E>increased risk for maternal and fetal complications</E> and may benefit from <A>early identification and treatment</A>.}
\end{example}

\subsection{Inter-annotator agreement for the medical data}
Due to the intricacy of causality extraction, which involves annotators labeling varying text spans as "cause," "effect," and so on, computing agreement between two annotators can be challenging as it requires comparing two spans of texts. Traditional methods of inter-annotator agreement, such as the Kappa statistic \cite{fleiss2013statistical}, are inadequate due to their need for classifications to fit into mutually exclusive and discrete categories. Therefore, we decided to assess agreement using both exact match and relaxed match criteria.
The F-measure is used for the exact match \cite{hripcsak2005agreement, thompson2009construction, mihuailua2013biocause} between the labels. 
In the case of the relaxed match, the average distance between phrases is computed. Initially, the annotated phrases, their corresponding labels, and the full sentence they are derived from are extracted from the entire annotated document. These annotations, originating from both annotators, are then compared and amalgamated based on the sentence. The resulting merged table thus features the sentence, the extracted phrase, and the labels as marked by Annotator 1 and Annotator 2. In total, 514 matching phrases have been identified. An overall agreement computed as a Jaccard similarity of 0.66 was obtained. Details of the inter-annotator agreement computation are given below. 

\FloatBarrier
\begin{figure}
\centering
\includegraphics[width=8cm,height=7cm]{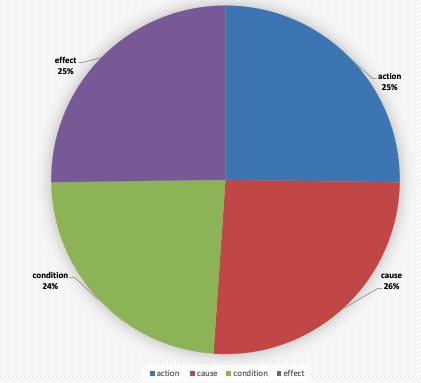}
\caption{Distribution of the labels in the corpus. The percentage of almost all the labels is around 24\%.}
\label{fig:label_count}
\end{figure}

From the merged data table, the inter-annotator agreement was computed. This is done by computing the match between the annotations as follows.
\begin{itemize}
    \item Relaxed match -- Both annotator's phrases overlap with each other but are not necessarily an exact match.
    \item Exact match -- Both annotator's phrases exactly match.
\end{itemize}
 
To execute the relaxed match, we employed the Levenshtein distance \cite{miller2009levenshtein} and the Jaccard distance \cite{real1996probabilistic}. The Levenshtein distance quantifies the difference between two string sequences, indicating the minimum single-character edits required to transform one word into another. Jaccard similarity computes the degree of relatedness between two finite samples by dividing the intersection's size by the size of the sample sets' union. The Jaccard distance is subsequently calculated by subtracting the Jaccard similarity from 1.
The Python library Levenshtein \footnote{\url{https://pypi.org/project/python-Levenshtein/}} is used in computing the Levenshtein distance. The Jaccard index was computed using the Python library textdistance\footnote{\url{https://pypi.org/project/textdistance/}}.
The Levenshtein distance and the Jaccard distance between the annotators are summarized in Table~\ref{tab:Inter-annotatoragreement}

\begin{table}[]
\centering
\begin{tabular}{|l|l|l|}
\cline{1-3}
      & Levenshtein distance & Jaccard distance \\ \hline
Cause & 0.22 & 0.27  \\ \hline
Condition & 0.34 & 0.21  \\ \hline
Effect & 0.37 & 0.31   \\ \hline
Action & 0.87 &  0.48  \\ \hline
\end{tabular}
 \caption{Relaxed match between the annotated phrases. Levenshtein distance is the minimum number of edits required to transform one phrase to another, whereas Jaccard distance is the amount of non-overlap between phrases. The lower the distance, the agreement is higher. The distance is higher for action. In most of the cases where there is a mismatch, the length of the phrase by both the annotators was different.}
    \label{tab:Inter-annotatoragreement}
\end{table}

From Table~\ref{tab:Inter-annotatoragreement}, we can understand that there is an average Levenshtein distance of 0.41 and an average Jaccard distance of 0.34. In most cases, both annotators annotated the same sentence with the same labels, but the length of the phrase was different. 
The exact match between the phrases is computed by finding the exact string match between phrases 1 and 2. Out of the 514 phrases, 112 phrases are exact matches.
The match between the labels for the same phrase by both annotators is also computed with an average F1 score of 0.78. The match between the labels for each subcategory is given in Table~\ref{tab:Matchbetweenlabels}

\begin{table}[]
\centering
\begin{tabular}{|l|l|l|l|}
\cline{1-4}
      & Precision & Recall & F1-score  \\ \hline
Cause & 0.86 & 0.71 & 0.77  \\ \hline
Condition & 0.56 & 0.85 & 0.67  \\ \hline
Effect &0.85 & 0.90 & 0.88  \\ \hline
Action & 0.89 & 0.70 & 0.78  \\ \hline
\end{tabular}
 \caption{For a given phrase, the labels annotated by annotators 1 and 2 are compared. An average F1 score of 0.78 was obtained. From the F1-score, we can understand that both the annotators agree on most of the categories except the signal for which the F1-score is low. }
    \label{tab:Matchbetweenlabels}
\end{table}

\subsection{Data preparation and preprocessing}
Seven documents on gestational diabetes guidelines provided by different societies are downloaded as PDF documents. The PDFs are converted into a document format, and the documents are given to the annotators for annotating them manually. The annotators used tags to annotate the documents. 

After annotating them, the NLTK sentence tokenizer is used to extract sentences from all the documents. The sentences from all the documents are appended together and converted into a data frame. Regular expressions are used to extract the causal sentence. If any of the sentences contain a tag <>, it will be extracted as a causal sentence. Again regular expressions are used to extract the phrases of cause, effect, action, signal, and condition from the sentences. The extracted phrases are used for computing inter-annotator agreement. 

\section{Methodology}

\subsection{Causality extraction using BERT}
Given the good performance of DistilBERT with organizational data \cite{gopalakrishnan2023text}, this model was also applied to the medical data. Considering the limited sample size in medical data, we attempted to improve the learning process by increasing the number of epochs. This approach allows for more refined fine-tuning of the model.

\begin{figure}
\centering
\includegraphics[width=10cm,height=8cm]{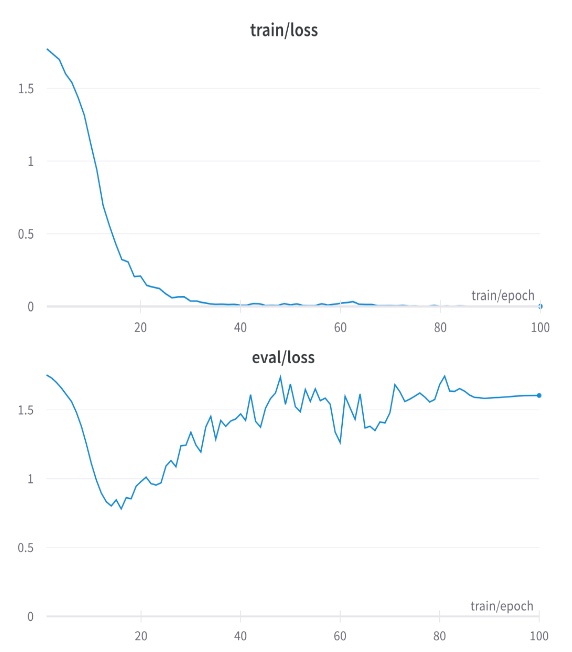}
\caption{Graph showing the train and validation loss when fine-tuning on BioBERT. Looking at the graph, we can understand that with the increase in the number of epochs, the training loss is constantly decreasing and approaching 0. The validation loss decreases till 16 epochs and then starts to increase. Based on this, we fine-tuned BioBERT for 16 epochs. }
\label{fig:BioBERT-combined}
\end{figure}

In order to decide on the correct number of epochs and to avoid overfitting, we tried running the model for 100 epochs and plotted the validation loss and the training loss. 
The graph showing the train and validation loss for our highest performing model, BioBERT, is given in Figure~\ref{fig:BioBERT-combined}.

From the graph, we can understand that with the increase in the number of epochs, the training loss is constantly increasing and approaching 0. The validation loss decreases till 18 epochs and then starts to increase. Based on this, we fine-tuned DistilBERT for 18 epochs, BERT(BERT-base-uncased) for 20 epochs, and BioBERT for 16 epochs. 

The data is split into train and test. DistilBERT for token classification is fine-tuned on the training data for 18 epochs. On the test data, the model obtained an average F1-score of 0.57. 
Similarly, we fine-tuned BioBERT for 16 epochs and BERT for 20 epochs. Out of these three models, BioBERT\cite{lee2020biobert} gave us an average higher F1-score. BioBERT gave an average F1 score of 0.61, and BERT gave an average F1 score of 0.60. The detailed results of fine-tuning BioBERT on the test data are given in Table~\ref{tab:BioBertcausalityextraction};
and, for comparison,
the summary of the results of using variants of BERT for causality extraction task is given in Table~\ref{tab:summarycausalityextraction}

\begin{table}[]
\centering
\begin{tabular}{|l|l|l|l|l|}
\cline{1-5}
 & Precision & Recall & F1-score & Support  \\ \hline
E & 0.82 & 0.75 & 0.78 & 696 \\ \hline
C & 0.62 & 0.69 & 0.65 & 411  \\ \hline
CO & 0.80 & 0.63 & 0.71 & 717   \\ \hline
A & 0.65 & 0.85 & 0.73 & 838   \\ \hline
Macro average & 0.72 & 0.73 & 0.72 & 2662   \\ \hline
\end{tabular}
 \caption{Causality extraction results on the medical data using BioBERT, the highest performing model. Each token in the text was assigned a label  Effect(E), Cause(C), Condition(CO), and Action(A). The results are obtained by splitting the manually annotated data into train and test data.}
    \label{tab:BioBertcausalityextraction}
\end{table}

\begin{table}[]
\centering
\begin{tabular}{|l|l|l|l|}
\cline{1-4}
 & Precision & Recall & F1-score   \\ \hline
DistilBERT & 0.69 & 0.68 & 0.68 \\ \hline
BERT & 0.72 & 0.72 & 0.71  \\ \hline
BioBERT & 0.72 & 0.73 & 0.72   \\ \hline
\end{tabular}
 \caption{Summary of the results of causality extraction on medical text using the Pre-trained Language Model (BERT) and its variants. The gestational diabetes data is split into train and test data. All the models are fine-tuned on train data and tested on test data.}
    \label{tab:summarycausalityextraction}
\end{table}

\subsection{Observations on using GPT-4 for causality extraction from medical guidelines} 
%
%
%
Generative Pre-trained Transformer 4 (GPT-4) \cite{openai2023gpt4} outperforms most of the state-of-the-art performing models on the traditional NLP benchmark datasets. 
In this section, we discuss our results of prompting GPT-4-0314 with a with context window of 8,192, for the causality extraction task. 
We explored various prompt sizes (zero, four, six, eight, ten-shot, and twenty-shot prompting). 

As an initial step, we tried the sentence with token-level labels for each word in the sentence as prompt examples. For the test data, the model is expected to predict a label for each word in the sentence. However, the model hallucinated by predicting a longer number of labels than in the given sentence; that is, a long sequences of non-existing "non-causal" labels. 

Since GPT-4 hallucinated for the token-level predictions, we tried extracting the phrases of cause/effect relationships in text and tried converting them into token level by assigning labels for each token. 
We started with a four-shot prompting. The annotated data with the tags will be given as an example in the prompt, and the model is expected to predict similarly. A sample is given in Example 4.1.
\begin{example}
    \texttt{<C>Gestational diabetes</C>  has also been associated with an <E>increased risk of several long-term health outcomes in pregnant persons and intermediate outcomes in their offspring</E>}
\end{example}

We tried converting the predictions with the tags into a token-level format in order to compute the F1 score. However, since the tags are placed in different places in some of the gold annotations and predictions, the number of tokens in gold and predictions doesn't match. An example is given below
\begin{example}
\texttt{
   \textbf{Gold:} Importance<C>Gestational diabetes</C>  is diabetes that develops during pregnancy.1-3 Prevalence of gestational diabetes in the US has been estimated at 5.8\% to 9.2\%, based on traditional diagnostic criteria, although it may be higher if more inclusive criteria are used.4-8 <C>Pregnant persons with gestational diabetes</C> <E>increased risk for maternal and fetal complications, including preeclampsia, fetal macrosomia (which can cause shoulder dystocia and birth injury), and \texttt{neonatal hypoglycemia</E> .3,9-11} <C>Gestational diabetes</C>  has also been associated with an <E>increased risk of several long-term health outcomes in pregnant persons \textit{and intermediate outcomes in their offspring</E> .12-16Table 1.}} \\

   \texttt{
   \textbf{Prediction:} Importance Gestational diabetes is diabetes that develops during pregnancy. 1-3 Prevalence of gestational diabetes in the US has been estimated at 5.8\% to 9.2\%, based on traditional diagnostic criteria, although it may be higher if more inclusive criteria are used.4-8 <C>Pregnant persons with gestational diabetes</C> <E>increased risk for maternal and fetal complications, including preeclampsia, fetal macrosomia (which can cause shoulder dystocia and birth injury), and \texttt{neonatal hypoglycemia. 3,9-11</E>} <C>Gestational diabetes</C> has also been associated with an <E>increased risk of several long-term health outcomes in pregnant persons \texttt{and intermediate outcomes in their offspring.12-16Table 1.</E>}}
\end{example}

In Example 4.2, the phrases marked indicate the scenario where some extra spaces can be added, leading to the indifference in the number of tokens between gold and the predictions. In the gold data, \texttt{neonatal hypoglycemia</E> .3,9-11 have a space after the tag, but in the prediction, the tag is predicted after \\
the number, which leads to no space between </E> and .3,9-11.} In some scenarios, the GPT-4 omits some of the words if they do not contain a causal relation (omits the 'O' labels in some places).  This mismatch between the gold and the predictions impedes the token-level comparison and reporting of the F1 score. An example is given below:
\begin{example}
\texttt{
\textbf{Gold:}Race/Ethnicity/Hemoglobinopathies<C>Hemoglobin variants </C>  can <E>interfere with the measurement of A1C</E>, although most assays in use in the U.S. are unaffected by the most common variants.}

\texttt{
\textbf{Prediction:} <C>Race/Ethnicity/Hemoglobinopathies variants</C> can interfere with the measurement of A1C, although most assays in use in the U.S. are unaffected by the most common variants.}
\end{example}
In Example 4.3, in the prediction, the keyword "Hemoglobin" is missing, which is present in the gold data. In some places, such inconsistencies lead to token mismatch between the gold and predicted data. 

To compare the performance of GPT-4 with other models, the predictions are converted into the token level and manually checked to convert both the gold predictions to the same number of tokens for the four-shot prompting. In the predictions, some tokens are missed; those tokens are added to the predictions and marked as label "O."(as O indicates tokens that are not cause, effect, condition, action, or signal). After converting the data into a token level, we computed the F1 score. With GPT-4, we got an average F1 score of 0.39 with four-shot prompting. 


\section{Results \& Experiments}
As the predictions of GPT-4 can be unreliable,  and missing tokens in a sentence leads to a token mismatch between the gold data and the predicted data, therefore Jaccard distance is proposed as an alternative solution to the traditional F1 score as the evaluation criteria. The Jaccard similarity was computed using the textdistance\footnote{\url{https://pypi.org/project/textdistance/}} Python library. 
Another alternative measure to try is the cosine similarity.
The cosine similarity is obtained by computing the vectors of both the gold and the predictions using the Universal Sentence Encoder\cite{cer2018universal}. The computed values are used to compute the pairwise cosine similarity between two vectors using Scikit-learn \footnote{\url{https://scikit-learn.org/stable/modules/generated/sklearn.metrics.pairwise.cosine_similarity.html}}.
The cause, effect, signal, condition, and action are extracted from the predictions using regular expressions on the tags. The extracted prediction phrases and the gold annotated phrases are merged. We perform two types of matching on the gold and predicted phrases. 
\begin{itemize}
    \item \textit{Jaccard similarity:} To measure the dissimilarity between the gold data and the predictions.

    \item \textit{Cosine similarity:} To measure the semantic similarity between the gold data and the predictions. 
\end{itemize}

The results of phrase level similarity between the gold annotated data and predictions of GPT-4 using various prompt sizes are summarized in Table~\ref{tab:GPT4-CE-result}.
\begin{table}[]
\centering
\begin{tabular}{llll}
 & Jaccard similarity & Cosine similarity & F1 (labels)  \\ \hline
Zero-shot & 0.42 & 0.22 & 0.27   \\ \hline
Four-shot & 0.44 & 0.22 & 0.35  \\ \hline
Six-shot &  0.57 & 0.22 & 0.52 \\ \hline
Eight-shot & 0.52 & 0.23 & 0.55 \\ \hline
Ten-shot & 0.57 & 0.20 & 0.60  \\ \hline
Twenty-shot & 0.46 & 0.20 & 0.28

\end{tabular}
\caption{The phrase level comparison results of few-shot prompting using GPT-4. We tried various prompt sizes (zero, four, six, eight, ten, and twenty-shot prompting). From the results, we can understand that the Jaccard similarity at ten-shot prompting is higher (higher the similarity, higher overlap between the gold and predicted spans), cosine similarity is lower (lower the similarity, higher the gold and press are related), and the F1-score between the labels is higher, after which the similarity and F1 decreases at twenty-shot. The cosine similarity, which gives the semantic similarity between gold and predictions, remains the same with all the prompt sizes. Here the F1-score is computed by comparing the gold labels and the predicted labels. 
} \label{tab:GPT4-CE-result}
\end{table}

From the results of the various prompt sizes for the causality extraction on medical data, we can understand that the result of the ten-shot prompting gives a higher similarity and F1 score.

 The Jaccard similarity gives the similarity score based on the overlap between the gold and the predictions.  The cosine similarity gives the semantic similarity between the gold and predicted phrases. There is not much difference in the cosine similarity with various prompt sizes, indicating that it may not be the right measure for this task. 
The F1-scores are computed by comparing the gold labels with the predicted labels (Jaccard and cosine similarity for the predicted phrases, F1-score for the labels).
The detailed F1-score for the ten-shot prompting label match between gold and predictions is given in Table~\ref{tab:GPT4-CE-F1score}. 
In particular, we can see that the F1 score for cause, effect, and action is higher compared to the other labels. 
(This result is comparable with a recent work \cite{chan2023chatgpt}, indicating a strong performance of ChatGPT for extracting cause/effect relationships).  

\begin{table}[]
\centering
\begin{tabular}{llll}
 & Precision & Recall & F1 score  \\ \hline
Action & 0.56 &  0.90 &  0.69  \\ \hline
Cause & 0.60 &  0.74 &  0.66  \\ \hline
Condition & 0.95 &  0.17 & 0.29   \\ \hline
Effect & 0.74 &  0.79 &  0.76  \\ \hline
Macro average& 0.71 &  0.65 &  0.60  \\ \hline
\end{tabular}
\caption{Summary of the results of the GPT-4 predictions of ten-shot prompting on our medical data. Here the F1-score is computed by comparing the gold labels and the predicted labels. The F1 score for cause, effect, and action is higher compared to the condition. Many of the conditions are predicted as causes. 
} \label{tab:GPT4-CE-F1score}
\end{table}

\subsection{LLAMA2 for causality extraction from medical guidelines}

LLAMA2\cite{touvron2023llama} is a pre-trained and fine-tuned Large Language Model. Three variants of LLAMA2 are available, which differ in the parameters. 7B, 13B, and 70B parameters are publicly available. LLAMA2 is trained on two trillion tokens of data. 
In our experiments, the LLAMA2 7B parameter is fine-tuned on the medical data. It is fine-tuned using the HuggingFace autotrain. 

To fine-tune LLAMA2, the first step is to prepare the data. At first, when the model was fine-tuned and tested on the token level as BERT, LLAMA2 was predicting a long number of "O-other" as GPT-4. So we dealt with this as a phrase-level extraction problem. The data is prepared with three parts which are instruction, input, and output. A sample training data is given in example 5.1

\begin{example}
\texttt{
    \#\#\#Instruction: Extract the cause, condition, effect, signal, and action from the given sentence.  \#\#\#Input: Pregnant persons with gestational diabetes are at increased risk for maternal and fetal complications, including preeclampsia, fetal macrosomia (which can cause shoulder dystocia and birth injury), and neonatal hypoglycemia. \#\#\#Output: ['Pregnant persons-signal', 'with gestational diabetes -cause', 'increased risk for maternal and fetal complications, including preeclampsia, fetal macrosomia (which can cause shoulder dystocia and birth injury), and neonatal hypoglycemia-effect']
}
\end{example}

The test data should be similar to the training data except for the output, which should be empty. The gestational diabetes annotated data was split into train and test data. The HuggingFace autotrain \footnote{\url{https://huggingface.co/docs/autotrain/llm_finetuning}} for the LLM fine-tuning was used to fine-tune the model. The fine-tuned weights are pushed into the HuggingFace dataset for inference. This experiment was done using Google Colab Pro+ with a High-RAM A100 GPU. Similar to the GPT-4, the predictions of LLAMA-2 were also at phrase level. So a similar evaluation strategy is followed for LLAMA2. We present the results with three types of distance. 

The predictions are split into phrase levels and then compared with gold data. 
The Jaccard similarity was computed using the textdistance\footnote{\url{https://pypi.org/project/textdistance/}} Python library. 
The cosine similarity is obtained by computing the vectors of both the gold and the predictions using the Universal sentence encoder\cite{cer2018universal}. The computed values are used to compute the pairwise cosine similarity between two vectors using Scikit-learn \footnote{\url{https://scikit-learn.org/stable/modules/generated/sklearn.metrics.pairwise.cosine_similarity.html}}. 

Initially, we split the data into train and test using the Scikit learn train\_test\_split(). We have converted the phrase-level predictions into token-level. In the test data, there were a total of 59 samples. Out of the 59 samples, only 29 samples, LLAMA2 predicted the labels, so the evaluation is only for those sentences. With LLAMA2, we got an average F1-score of 0.36, which is lower than that of all the other models.  

Since the test data size is very small, we have also tried a four-fold cross-validation on this data. The results of fine-tuning LLAMA2 using four-fold cross-validation with 3,5, and 10 epochs are given in Table~\ref{tab:LLM-result}. 

With the increase in the number of epochs, both the Jaccard similarity and F1-score increase. 
Also, the predictions of LLAMA2 missed labels in many of the predictions. It extracted the phrases with no label. With three epochs, LLAMA2 missed 38\% of the labels; with five epochs, 21\% of the labels; and with ten epochs, it missed 26\% of the labels. We omitted the predictions with no labels (108 predictions, 60 predictions, 76 predictions). The results of causality extraction presented in Table~\ref{tab:LLM-result} are after omitting the predictions with no labels. 

From the results, we can understand that Jaccard similarity, cosine similarity, and F1 score increase with the increase in the number of epochs. 
However, the number of missed labels started increasing after 5 epochs. 






\begin{table}[]
\centering
\begin{tabular}{llll}
 & Jaccard similarity & Cosine similarity & F1-score\\ \hline
LLAMA2 (3epochs) & 0.73 & 0.19  &  0.70  \\ \hline
LLAMA2 (5epochs) & 0.888 & 0.20 & 0.75 \\ \hline
LLAMA2 (10epochs) & 0.90 & 0.21 & 0.76 \\ \hline
\end{tabular}
\caption{The phrase level comparison results of LLAMA2 using 4 fold cross validation. Jaccard similarity and cosine similarity indicate the average similarity between the gold and the predictions. The F1 score is the comparison between the gold labels and predicted labels. 
} \label{tab:LLM-result}
\end{table}


\section{Discussion}

Above we presented results on causality extraction from medical guidelines using recently introduced large language models such as LLAMA2 and GPT-4, and compared them with the performance of BERT, an older, and smaller LLM. 
The annotated data and the code are all publicly available on GitHub: \url{https://github.com/gseetha04/LLMs-Medicaldata.git}
.

We observed that GPT-4 expresses strong performance for the cause-effect relationships with medical data, and generally has a good understanding of medical text without fine-tuning. However, in contrast with GPT-3.5 
 GPT-4 cannot deal with token classification, which limits the traditional way of finding cause and effect phrases, as discussed e.g. in our previous work \cite{gopalakrishnan2023text}. 
 
Even though LLAMA2 seems to perform well for causality extraction, the predictions of the LLAMA2 do not predict labels for many cases, 
which limits its practical application. This perhaps was caused by fine-tuning LLAMA2 on our small dataset. Therefore, increasing the size of the dataset before the fine-tuning may improve the performance. However, large annotated datasets for CPGs are not available, and further experiments would require annotating more data. 
Since we focused on the accuracy of actual predictions, we omitted 38\% of labels with three epochs, 21\% of labels with five epochs, and 26\% of labels with ten epochs. 

Given its relatively high performance and ease of use, BERT-based models continue  to be a state-of-the-art for causality extraction tasks,  even in the age of LLM . 

\section{Conclusion}
We developed an automated technique for extracting causalities from annotated corpora of medical guidelines. Additionally, we exhibited the practicality of employing new Large Language Models for causality extraction tasks. With BioBERT, we got an average F1-score of 0.72, whereas with LLAMA2, an average Jaccard distance of 0.40 was obtained. We demonstrated the potential for extracting causalities from medical guidelines using a small annotated corpus. The next logical step could involve expanding the corpus through the annotation of more data and creating a benchmark dataset for causality extraction from medical guidelines.

The potential of this research opens up novel dimensions for the health domain, as causality extraction from medical guidelines can enhance clinical decision-making and patient care. This work explored both machine learning and natural language processing techniques for causality extraction. Despite the abundance of causal sentences within these guidelines, automatic extraction is an unexplored field of research. Also, machine learning models often fail in clinical applications \cite{schmidt2017md} due to the gap between data (both training and testing). In order to avoid this gap, more realistic tests need to be done so that they can be employed for real-world data. 

\section*{Credit authorship contribution statement}\label{}

{
S.G. performed the majority of the experiments and writing. She also supervised the annotation process. L.G. chose the clinical guidelines data for annotations and participated in discussions and writing.  W.Z. designed some of the experiments, provided feedback, and contributed to writing.}

\begin{acks}
This research was partly funded by the National Science Foundation (NSF)  grant number 2141124. 
We would like to thank Nikhil Vundela for his contributions to data annotation. We would like to thank Dr. Wenwen Dou and Dr. Victor Zitian Chen for their feedback on the causality extraction tasks. 
\end{acks}

\bibliographystyle{ACM-Reference-Format}
\bibliography{sample-base}

\appendix

\end{document}